\documentclass[10pt,twocolumn,letterpaper]{article}

\usepackage[pagenumbers]{cvpr} 

\usepackage{amsfonts}       

\usepackage{multirow}

\usepackage{graphicx}
\usepackage{amsmath}
\usepackage{amssymb}

\usepackage{booktabs}
\usepackage[pagebackref,breaklinks,colorlinks]{hyperref}

\usepackage[capitalize]{cleveref}
\crefname{section}{Sec.}{Secs.}
\Crefname{section}{Section}{Sections}
\Crefname{table}{Table}{Tables}
\crefname{table}{Tab.}{Tabs.}

\def\cvprPaperID{7196} 
\def\confName{CVPR}
\def\confYear{2023}

\graphicspath{ {./resource/} }

\title{Primitive Graph Learning for Unified Vector Mapping}

\date{\today}

\author{Lei Wang$^1$,  Min Dai$^1$,  Jianan He$^1$,  Jingwei Huang$^1$,  Mingwei Sun$^{1,2}$ \\
	$^1$Riemann Lab, Huawei Technologies. $^2$Wuhan University \\
	\texttt{\{leiwang4, daimin11, hejianan2, huangjingwei6, sunmingwei2\}@huawei.com}\\
}

\hypersetup{
pdftitle={Primitive Graph Learning for Unified Vector Mapping},
pdfsubject={cs.CV, cs.LG, eess.IV},
pdfauthor={Lei Wang},
pdfkeywords={vector mapping, shape regularization, topology reconstruction},
}

\usepackage[skip=5pt]{caption}

\begin{document}
\maketitle

\begin{abstract}
Large-scale vector mapping is the foundation for transportation and urban planning. Most existing mapping methods are tailored to one specific type of ground target, due to the different shape topologies and shape regularization of different targets. We propose GraphMapper, an unified framework for end-to-end vector map extraction from satellite images.
Our key idea is an unified representation of vector maps of different topologies using a primitive graph, including a set of shape primitives and their pairwise relationship matrix.
%
Based on primitive graph, we design a learning approach to reconstruct primitive graphs in multiple stages. Through incremental primitive learning and pairwise relationship reconstruction, GraphMapper can fully explore primitive-wise and pairwise information for improved primitive graph learning. 
Additionally, our model achieves powerful context-aware shape regularization by learning directions that are consistent with pairwise relationship estimation. 
We empirically demonstrate the effectiveness of GraphMapper on two challenging mapping tasks for building footprints and road networks.
With the premise of sharing the major design of the architecture, and few task-specific designs, our model outperforms state-of-the-art methods in both tasks on public benchmarks.
Our code will be publicly available.
\end{abstract}

\section{Introduction}
Up-to-date vector maps are essential for navigation and urban planning. Methods to automatically extract vector maps from aerial or satellite images have greatly advanced in recent years. However, state-of-the-art vector mapping methods are tailored for one specific type of target. Consequently, multiple models must be maintained for comprehensive mapping systems, which increases the burden of model development and limits the system extensibility. 

The main challenge of designing an unified method for multiple vector mapping tasks lies in the difficulty to process different geometric primitives and topologies in an unified way. 
Therefore, 
specific designs have been required for different types of geometric primitives, such as point and line segment, for best performance \cite{Zorzi_2022_CVPR, He_2020_ECCV, Nauata_2020_ECCV, Girard_2021_CVPR}.
Additionally, accurate location and topology are the two basic requirements for vector mapping. Existing end-to-end methods either only focus on refining location accuracy \cite{Liang_2020_CVPR}, or try to perform feature learning on both tasks in parallel \cite{Zorzi_2022_CVPR,Zhang_2020_CVPR}. We argue that location and topology should be modeled incrementally, based on the intuition that improving location accuracy in advance can disambiguate and simplify topology learning.
%

The other challenge of high-performance unified mapping is that for certain mapping tasks, such as building mapping, it is critical to achieve aesthetic shape regularization beyond accurate geo-location and topology reconstruction. 
Shape regularization in vector mapping is essentially reducing the variation of relative relationship between primitives; for example, the angles between line segments of a building polygon usually share a few distinctive values (i.e, 0\textdegree, 90\textdegree, etc.).
By far, most methods regularize shapes through contour optimization or deformation \cite{Girard_2021_CVPR, Marcos_2018_CVPR, Cheng_2019_CVPR,CHEN2020114, Zorzi_2022_CVPR}. However, these methods rely on regressed information, which cannot explicitly enforce low variation of primitive relationships. Meanwhile, naively aligning shapes according to relationship types is sensitive to the errors of classified relationships, which further challenges effective learning-based shape regularization.
%

\begin{figure*}[t]
\begin{subfigure}{\textwidth}
	\includegraphics[width=0.9\textwidth]{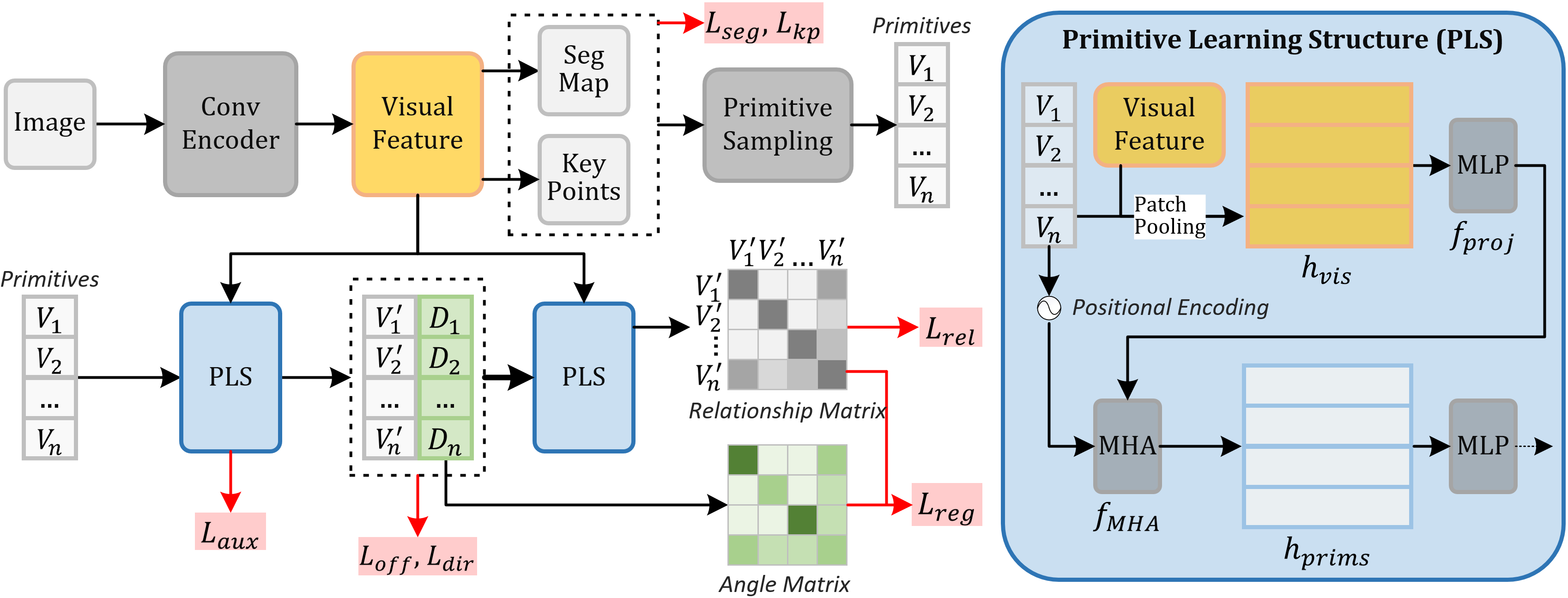}
	\centering
	\caption{GraphMapper workflow.}
\end{subfigure}
\begin{subfigure}{0.95\textwidth}
	\includegraphics[width=0.9\textwidth]{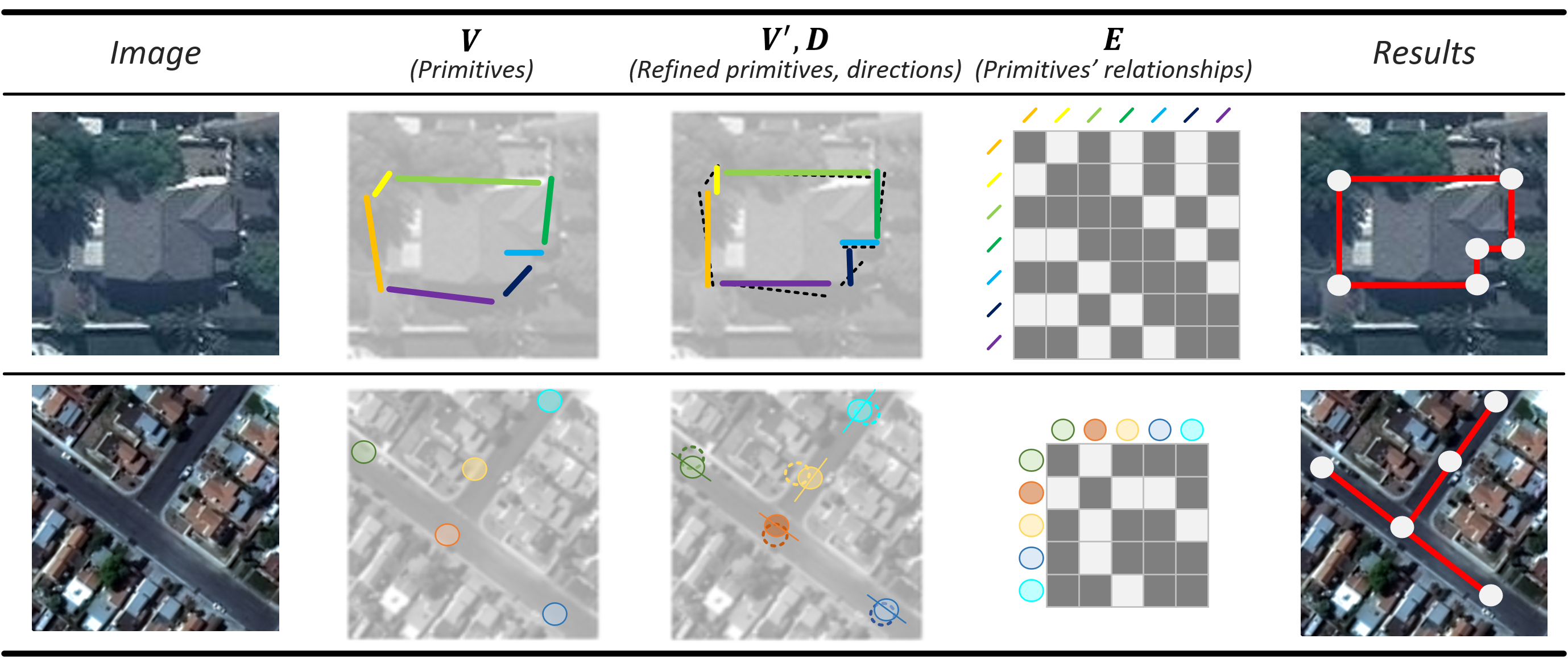}
	\centering
	\caption{GraphMapper applied to building and road mapping.}
\end{subfigure}
\caption{Overall workflow of GraphMapper. (a): It first segments input images and samples initial primitives from segmentation maps; then the initial primitives are refined by an primitive learning structure (PLS) which also predicts the direction of initial primitives; pairwise relationship between refined primitives are reconstructed using another PLS. (b): Illustration of GarphMapper applied to building and road mapping.}
\label{fig:workflow}
\end{figure*}

In this paper, we use primitive graph as a generic representation to build an \textit{unified framework}, GraphMapper, for multi-type vector mapping. GraphMapper incrementally learns to refine primitives' locations and reconstruct their pairwise relationships. Effective topology reconstruction and shape regularization are achieved through the relationship classification of pairwise primitives. With our design, most existing mapping tasks can be converted to image-based primitive graph reconstruction tasks. 
As shown in Fig. \ref{fig:workflow}, GraphMapper is mainly composed of a convolutional visual feature encoder and two primitive learning structures (PLS) using multi-head attention (MHA) network.
We first extract visual features of input image using a convolutional encoder and sample primitives (i.e., points, line segments) from the segmentation maps and keypoints. Then, we refine the primitives' coordinates and predict their geometric directions, which uses a PLS for local and global shape context modeling. Finally, we use another PLS to classify the relationships of primitive pairs. 


We apply GraphMapper to two typical mapping tasks: road and building mapping. For road, we predict point primitives and reconstruct road topology by classifying their pairwise connectivity relationship. For building, we find line segment primitives and their topology from segmentation results, and predict inlineness between line segments;
the network learns shape regularization by enforcing consistency between the network predicted angle matrix and pairwise relationship matrix of line segments.

By sharing most network components with few task-specific designs, GraphMapper outperforms state-of-the-art methods in both tasks in public benchmarks. In summary, our main contributions are the following:
\begin{itemize}
    \item We propose an unified learning framework based on primitive graph, which can be trained end-to-end for multi-type vector mapping tasks. 
    \item We perform the learning of location improvement and pairwise relationship classification incrementally, which effectively facilitates relationship reconstruction. 
    \item We use a relationship classification module to explicitly enforce shape regularization, in which primitives' directions and their predicted relationships are constrained to be consistent for robust performance.
\end{itemize}

\section{Related Works}
\label{sec:relatedwork}
\textbf{Unified vector mapping.} To the best of our knowledge, PolyMapper \cite{Li_2019_ICCV} is currently the only unified learning method for vector mapping. PolyMapper uses polygons as the unified representation for buildings and roads. A CNN-RNN structure is used to recurrently predict point sequences. However, representing road as polygons leads to redundant points, and predicting point sequence can easily introduce geometric errors. In contrast, GraphMapper supports learning on different primitives, and holistically reconstructs the topology of sampled primitives.

\textbf{Road mapping.} Road mapping methods focus on improving topological correctness for navigation purposes. 
Various techniques are made to improve the connectivity of road segmentation maps \cite{Mnih_2010, Zhou_2018_CVPR_Workshops,Mosinska_2018_CVPR, Batra_2019_CVPR,Batra_2019_CVPR,Etten_2020_WACV}. To reconstruct road topology from imperfect segmentation results, \cite{Mattyus_2017_ICCV} uses a binary decision classifier to predict the correctness of connections of nearby road endpoints from their image features, but it does not explore global road structure or shape prior. Graph-based methods \cite{Bastani_2018_CVPR, Tan_2020_CVPR, Li_2019_ICCV} reconstruct road networks by iterative searching of the next point in the road graph, using a CNN or CNN-RNN structure. Shape prior and global road structure are implicitly modeled in the iterative searching process. Compared to graph-based methods, our method generates a road graph in one forward run with all road points available, which allows easier integration of global and local shape contextual information.

Recently, Sat2Graph \cite{He_2020_ECCV} achieved the state-of-the-art performance by connecting segmented key points along road direction with carefully designed searching rules. Uniformly distributed road points and their directions are predicted using a multi-task CNN. Several steps of post-processing are performed to reduce artifacts and false connections. 
Compared to Sat2Graph, our method learns connectivity end-to-end, without complicated post-processing.

\textbf{Building mapping.}
Building mapping methods are now focusing on learning vector results from image inputs, where regularized shape representation is often a major concern.
Conventional methods often rely on heuristic rules \cite{MSBuilding, sirko2021continentalscale} for shape regularization, which is limited to simple scenarios and requires extensive tuning. 
PolygonRNN and its extensions \cite{Castrejon_2017_CVPR, Acuna_2018_CVPR, Li_2019_ICCV} achieved end-to-end vector mapping from images by recurrently predicting point sequences. However, RNN-based methods often fail to produce regularized shapes due to the lack of shape optimization design.

\cite{Marcos_2018_CVPR,Cheng_2019_CVPR,Girard_2021_CVPR} try to optimize contours using ACM (Active Contour Model) \cite{Kass_1988} guided by learned semantic information. Specifically, DSAC \cite{Marcos_2018_CVPR} and DARNNet \cite{Cheng_2019_CVPR} regress the weights of the contour smoothness term based on semantic information for shape generalization, which tend to generate under-regularized shapes due to inadequate design in simplified representation; \cite{Girard_2021_CVPR} further optimizes boundary skeletons to align with learned frame fields, but still requires complicated post-processing for shape regularization. 
Instead, PolygonCNN \cite{CHEN2020114} and Polygon Transformer \cite{Liang_2020_CVPR} predict point deformation of segmented contours, in which the results are regularized implicitly by improving the accuracy of building vertices; comparatively, our approach goes one step further by explicitly modeling shape regularity.
Recently, PolyWorld \cite{Zorzi_2022_CVPR} achieved great performance by first deforming keypoints' coordinates, and then reconstructing topology through classifying pairwise point connectivity relationship. Despite that GraphMapper follows a similar structure, the relationship classification module in our framework serves multiple purposes: topology reconstruction for road mapping, and explicit shape regularization for building mapping.

\cite{Zorzi_2020_ICPR} tried to regularize building footprint segmentation mask using a GAN loss, which can be considered as a parallel design to our geometry-based regularization method.

\section{GraphMapper}
Our main idea is to turn various vector mapping problems into an unified primitive graph estimation problem. In the following sections, We will first introduce primitive graph in Section \ref{primitive_graph}. Then we explain the design of GraphMapper's network architecture in Section \ref{network}, and training targets in Section \ref{train}. Lastly, we provide details of applying GraphMapper to road and building mapping in Section \ref{section:apply}.



\subsection{Primitive Graph} \label{primitive_graph}
\begin{figure}[t]
	\includegraphics[width=0.4\textwidth]{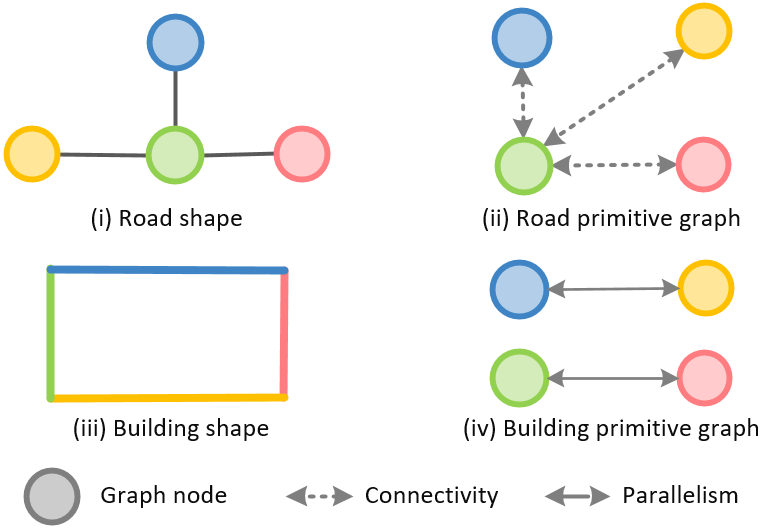}
	\centering
	\caption{Example primitive graph representations of a road junction and a building polygon.}
	\label{fig:primitive_graph_concept}
\end{figure}

A primitive graph is a homogeneous un-directed graph:
\begin{equation}
    G=\{V, E\}, V\in \mathbb{R}^{N \times d}, E\in \mathbb{Z^+}^{N\times N}
\end{equation}
where $V$ represents $N$ primitives with $d$-dimensional coordinates. A point primitive is represented by its image coordinate $(x, y)$; a line segment primitive is represented by the coordinates of its two end points $(x_1,y_1,x_2,y_2)$. $E$ is the relationship matrix, in which $E_{ij}$ represents the relationship between $V_i$ and $V_j$. Example primitive graph representation for road and building are shown in Fig. \ref{fig:primitive_graph_concept}. Depends on the choice of primitives and pairwise relationships, primitive graph can model various types of targets.

\subsection{Network Architecture} \label{network}
The overall structure of GraphMapper is illustrated in Fig. \ref{fig:workflow}. We reconstruct primitive graphs in three sequential steps. We first extract initial primitives from input images. Then, the location of initial primitives are refined using a primitive learning structure (PLS). The direction or normal direction of each primitive is also estimated at this stage. Lastly, we reconstruct the relationship matrix between pairwise primitives using another PLS.

\subsubsection{Primitive Detection} \label{primitive_detection_network}
We detect primitives by sampling from semantic maps predicted from input images. Previous methods tried to sample points from keypoint heatmap \cite{He_2020_ECCV, Zorzi_2022_CVPR}. We found this method often miss-detect points. Therefore, we sample additional keypoints from semantic segmentation maps.

More specifically, we use a FPN network with resnet backbone to encode input image $I\in\mathbb{R}_{+}^{3\times H\times W}$ into an multi-scale feature $F$. Then, we predict a semantic segmentation map $Y_{seg}\in\mathbb{R}_{+}^{S \times H\times W}$ and a keypoint heatmap $Y_{kp}\in\mathbb{R}_{+}^{K \times H\times W}$ from $F$ using two FCN heads. 

To sample point primitives, we first extract local maximum points using NMS on all non-background classes of $Y_{seg}$ and $Y_{kp}$, resulting $S+K-2$  of candidate points.
Then, we combine all candidate points and apply another NMS with category-specific priority scores to remove redundant points to get the sampled points $V'\in\mathbb{R}_{+}^{N\times 2}$, so that points of higher priority categories are kept over lower ones.

We sample line segments by connecting sampled key points. For polygon structures (i.e., building, forest), we trace contours from segmentation maps, which are then simplified using the Douglas–Peucker (DP) algorithm \cite{douglas_1973}. Points of the simplified contours are combined with sampled key points. We project points to their nearest contours, and connect points according to their projections' sequence in contour \cite{CHEN2020114}. Note that when the targets are represented by polygons, this method can accurately derive the relative sequence between line segments without learning the connectivity between line segments.

\subsubsection{Primitive Graph Reconstruction}
Given the initialized primitives as input, we first refine the coordinates of primitives to improve location accuracy and alleviate the difficulty of relationship learning. Then, we predict the pairwise relationship of the refined primitives. As both tasks benefit from shape context information, they share a common primitive learning structure. Different from previous methods \cite{Zorzi_2022_CVPR}, we reconstruct pairwise relationship based on refined primitives instead of initial primitives. Using refined primitives can reduce the ambiguity for relationship recognition and ground truth relationship generation, which is essential for high quality primitive relationship reconstruction.

\textbf{Primitive Learning Structure.} The structure of Primitive Learning Structure (PLS) is illustrated in Fig. \ref{fig:workflow}. Given primitives $V$ and their visual features as input, visual features at primitives' locations are pooled using patch pooling. Patch pooling extracts a small crop of image feature centered at each primitive, and compresses the cropped features using a small FCN network. The resulted primitive features $h_{vis}$ are flattened and individually projected using a MLP ($f_{proj}$), the result of which is fed into a multi-layer MHA module $f_{MHA}$ \cite{Vaswani_2017_NIPS} together with the positional encoding of primitive coordinates \cite{carion_2020_ECCV}. MHA can fuse geometric and visual information and exchange information among primitives to generate local and global contexturized primitive features $h_{prims}$, which are used by MLP heads to generate output predictions.

\textbf{Primitive Refinement.} We use a PLS with two MLP heads to predict the coordinate deformation and directions $D\in \mathbb{R}^{N \times 2}$ (normal direction for points and line direction for line segments) of input primitives from PLS generated primitive feature $h_{prims}$. We get refined primitives $V'$ by adding estimated deformation back to input coordinates. Note that line segments' direction can also be computed from their coordinates $V'$. However, we found network predicted $D$ is more accurate, as it is easier to regress a direction than adjusting both vertices of a line segment to achieve the desired angle while staying close to segmentation boundary.

For direction regression, the discontinuity of rotation angles at 0 and 180° can lead to unstable learning \cite{Zhou_2019_CVPR} when naively applying \emph{L2 loss} on direction angles $D$. Therefore, we regress the sine and cosine of a surrogate angle which is 2 times the target angle $A$ as suggested in \cite{Zhou_2019_CVPR}:
\begin{equation}
    \overline{D}_i = (cos(2A_i), sin(2A_i)).
\end{equation}
Each row in $D$ is a normalized 2-dimensional vector. During inference, we recover the actual direction $D'$ from surrogate direction $D$.

\textbf{Primitive Relationship Reconstruction.} We input the refined primitives $V'$ and visual feature into another PLS to predict relationship matrix $E$. For a pair of point primitives, we extract additional visual features on the line segment between them using LOI \cite{Zhou_2019_ICCV} from visual feature, $Y_{seg}$ and $Y_{kp}$. These extracted features are concatenated together with their point features in $h_{prims}$ to form the point pair's feature. For a pair of line segment primitives, we simply concatenate their features in $h_{prim}$ as the pair's feature. For $N$ primitives, we get a relationship feature matrix $Q\in \mathbb{R}^{d_r\times N\times N}$. The MLP heads in PLS independently classify each pair's relationship using their feature in $Q$.

As only spatially neighboring primitives have positive relationship, the ratio of positive and negative relationship in relationship matrix $E$ is often strongly biased. Therefore, only the relationship of primitive pairs within a spatial distance threshold $t$ are used for training to balance positive and negative sample ratio.

\textbf{Shape regularization.} Primitive graph allows explicit representation of shape regularization as the consistency between primitives $V$ and their relationship matrix $E$:
\begin{equation}\label{eq:reg}
     L_{reg} = \sum_{i,j,r}E_{ijr}  |f_{prop}(V_i, V_j, r) - f_{prop}(\overline V_i, \overline V_j, r)|
\end{equation}
, where $f_{prop}$ computes the desired property between primitives. The term in $|\cdot|$ represents the difference between computed and desired property between $V_i$ and $V_j$ for relationship category $r$. $\overline V_i$ and $\overline V_j$ are the corresponding ground truth primitives for $V_i$ and $V_j$. 
This formulation explicitly enforce low variation of primitives' relative properties. Additionally, by adjusting the strength of regularization according to the probability of its relationship type, the network can learn to balance between shape regularity and location accuracy.

\subsection{GraphMapper Learning} \label{train}
GraphMapper is trained with a linear combination of the following losses:
\begin{equation}
    (\mathcal{L}_{seg}, \mathcal{L}_{kp}, \mathcal{L}_{off}, \mathcal{L}_{dir}, \mathcal{L}_{ret}, \mathcal{L}_{reg}, \mathcal{L}_{aux})
\end{equation}
, which we explain below.

\textbf{Segmentation loss $\mathcal{L}_{seg}, \mathcal{L}_{kp}$.} $L_{seg}$ is a linear combination of cross-entropy loss and Lovász-softmax loss \cite{berman2018lovasz} applied to semantic segmentation map $Y_{seg}$. Lovász-softmax loss is similar to Dice loss \cite{sudre_2017} that penalizes the structure of segmentation maps, but achieves more stable training. $L_{kp}$ is a cross-entropy loss applied to key point segmentation map $Y_{kp}$.

\textbf{Primitive deformation loss $\mathcal{L}_{off}$.}
$L_{off}$ is a bi-projection loss \cite{CHEN2020114} that penalizes the deviation of the deformed primitives $V'$ from the ground truth primitives $\overline{V}$. It first matches the vertices in ground truth to its nearest predictions, then matches the rest of the predicted vertices to its nearest projection in ground truth shape. The mean \emph{L2} distance between matches is reported. 

\textbf{Primitive direction loss $\mathcal{L}_{dir}$.} $\mathcal{L}_{dir}$ is a loss applied to normalized surrogate direction $D$:
\begin{equation}
	\mathcal{L}_{dir} =  \frac{1}{|D|}\sum(D_i - \overline{D}_i)^2
	\label{eq:loss_dir}
\end{equation}
where $\overline{D_i}$ is the unit direction vector of 2 times the ground truth angle for primitive $V'_i$.

\textbf{Relationship loss $\mathcal{L}_{rel}$.}
$\mathcal{L}_{rel}$ is a cross-entropy loss applied to elements in $E$:
\begin{equation}
	\mathcal{L}_{rel} = -\frac{1}{|U|}\sum_{(i,j) \in U} E_{i,j}log(\overline{E}_{i,j}),
	\label{eq:loss_vrel}
\end{equation}
where $\overline{E}_{i,j}$ is the ground truth relationship matrix between primitives ${V'_i, V'_j}$) represented in one-hot format. $U$ is the set of primitive pairs that has distance less than $t$.  

\textbf{Shape regularization loss $\mathcal{L}_{reg}$.} We enable shape regularization by training with $L_{reg}$. We set $f_{prob}$ in Eq. \ref{eq:reg} to
\begin{equation}\label{eq:fprop}
f_{prop}(D_i,D_j, r) = \left\{\begin{matrix}
 \cos(2[{D}_{i}-{D}_{j}+2\pi)//\pi]) &, r=1 \\  
 0 &, r=0
\end{matrix}\right.
\end{equation}
, which computes the cosine between two line directions. Here $r=1$ means two primitives have certain relationship with fixed angles, such as parallel or perpendicular. $r=0$ means no relationship, therefore no regularization. Practically, we compute an angle matrix for all line segment pairs as shown in Fig. \ref{fig:workflow}. $L_{reg}$ is evaluated using the angle matrix and relationship matrix.    

\textbf{Auxiliary loss $\mathcal{L}_{aux}$.} To facilitate the learning of geometric shape features in MHA, we would like primitives' visual feature $h_{vis}$ to be more related with geometric properties. Therefore, we add an auxiliary predictor on features pooled by patch pooling in primitive refinement PLS to predict coordinate offsets and primitive directions. Hence, $\mathcal{L}_{aux}$ is a linear combination of a deformation loss and a direction loss applied to the auxiliary predictions.

\subsection{Implementation Details} \label{section:apply}
We only provide essential details here due to the limitation of space. Please refer to Supplementary for more implementation details.

\begin{figure*}[t]
	\includegraphics[width=0.9\textwidth]{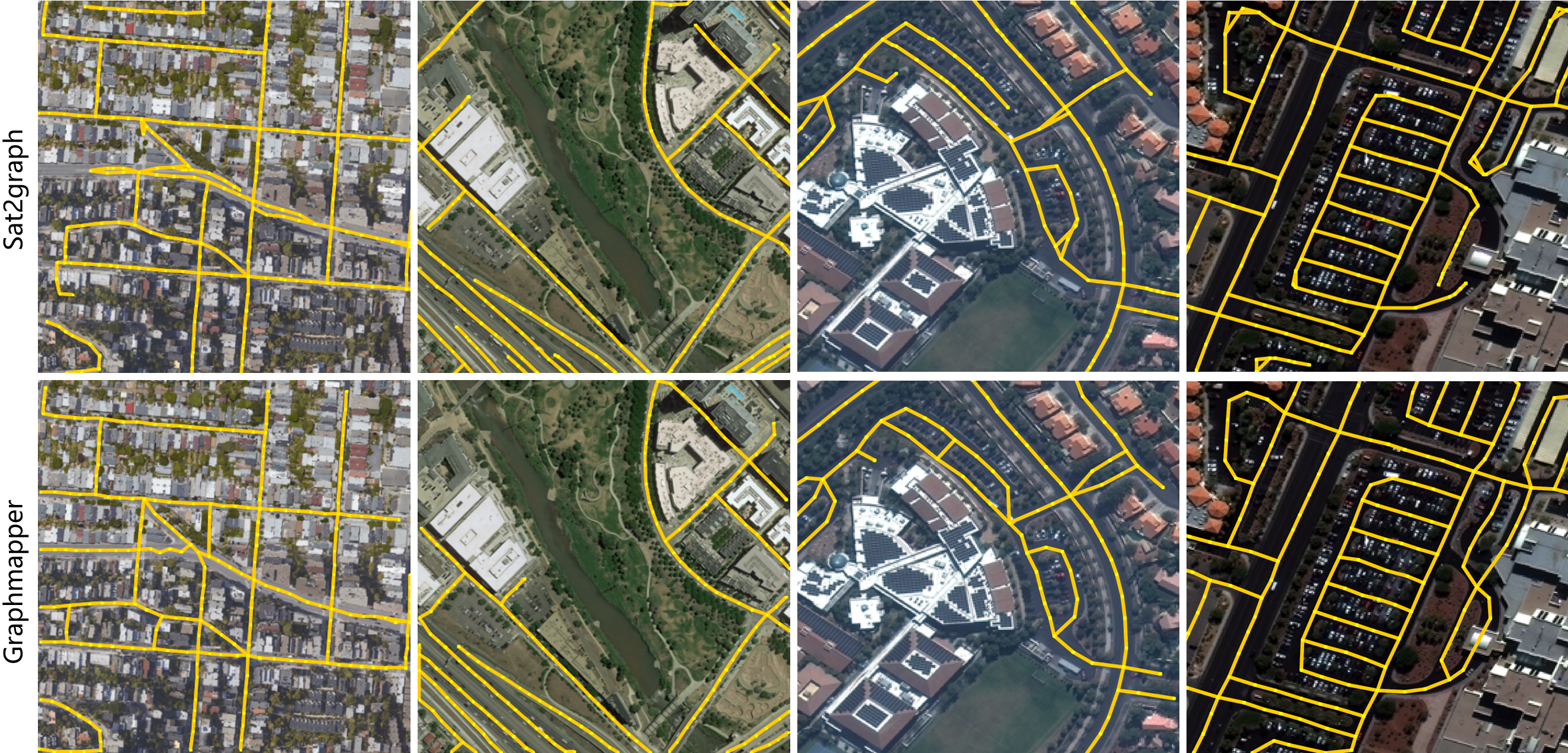}
	\centering
	\caption{Qualitative result of road mapping. Column 1-2 are from City-Scale dataset; column 3-4 are from SpaceNet Road dataset.}
	\label{fig:road_main}
\end{figure*}

\textbf{Training and testing.} To provide reasonable shapes for primitive graph reconstruction modules, we pre-train primitive detection before training all modules end-to-end. We use Adam optimizer \cite{adam} with batch size 12 and initial learning rate 2e-4. The max number of primitives per image is set to 150 for training and 300 for inference. Extra sampled primitives are discarded. 

\textbf{Road Network Mapping.} \label{section:apply_road}
We predict point primitives and pairwise connectivity for road mapping. We segment the buffered region of 5 pixels wide around road centerlines. Following \cite{He_2020_ECCV}, we predict four classes of key points in $Y_{kp}$: junctions, overlays (crossing point of overlapping road), end points (end points of road line segments that are not junctions or overlays), and interpolated points (points of fixed intervals interpolated on road line segments). 
Here, we don't use $\mathcal{L}_{reg}$ as primitive refinement can already achieve required shape regularization on road.
To reconstruct road network, we connect point $V'_i$ to a maximum of $t_i$ points that have connectivity probability in $E$ larger than threshold $t_r$. We set $t_i=3$ for junctions, and 2 for other points. We apply \emph{L2 normalization} to the primitive features before the last MLP layer of the second PLS to improve feature embedding quality, as inspired by contrastive learning studies \cite{NEURIPS2020_f3ada80d,Bardes2021,Chen_2021_CVPR}.

\textbf{Building Mapping.} \label{section:apply_building}
We predict line segment primitives and pairwise inlineness for building mapping. We trace contours from $Y_{seg}$, and reconstruct primitive graph for each group of line segments that belongs to the same contour, so that network can learn the context information within a building instance without the interference of other building geometries. At inference time, inline line segments are merged to simplify output polygons. Please see more details in Supplementary.
\section{Experiments}\label{sec:exp}

\subsection{Datasets And Metrics}
\textbf{Road Network.} (1) SpaceNet road dataset \cite{vanetten2019spacenet}: it contains 2549 satellite images of size 1300 $\times$ 1300 pixels with resolution around 0.3m. This dataset is challenging due to the diverse scenarios from 5 cities around the globe. Following the setting of \cite{He_2020_ECCV}, images are resized to 1 meter resolution, and the same train-test-valid splits are used. (2) City-Scale Dataset \cite{He_2020_ECCV}: it contains 180 tiles of size 2000 $\times$ 2000 with 1 meter spatial resolution. Ground truth vector annotations were collected from OpenStreetMap \cite{Mordechai_2008}. This dataset covers 20 U.S. cities, but with less diversity compared to the SpaceNet road dataset.

Road network topology is evaluated using TOPO \cite{Biagioni_2012_TRR} and Average Path Length Similarity (APLS) \cite{vanetten2019spacenet}. TOPO measures the similarity of sub-graphs near seed points sampled from the inferred graph and ground truth graph. The similarity of two matched graphs are evaluated as precision, recall and F1 score. The similarity of all sub-graphs are averaged and reported. APLS measures the similarity of graphs using the shortest path between sampled point pairs on each graph, which is more sensitive to topology structure compared to TOPO.

\textbf{Building.} CrowdAI Mapping Challenge Dataset \cite{mohanty2020deep} (CrowdAI dataset): It contains 280741 annotated aerial images for training and 60317 for testing. Each image has size 300 $\times$ 300 pixels.

We use IoU (Intersection Over Union) and AP/AR (Average Precision/Average Recall) to evaluate the overall correctness of generated polygons. Since IoU and AP/AR cannot describe the cleanness of predictions at boundaries, we adopt Mean Max Tangent Angle Error (MTE) \cite{Girard_2021_CVPR} and C-IoU \cite{Zorzi_2022_CVPR} as additional evaluation metrics to IoU and AP/AR.
MTE computes the max angle error of all line segments for each building and report the average value over the entire dataset \cite{Girard_2021_CVPR}. C-IoU is IoU weighted by polygon simplicity, where more points in predicted polygon means lower polygon simplicity and lower C-IoU. Here, polygon simplicity is evaluated using \emph{N ratio}, which is the ratio of the number of vertices between predictions and ground truth \cite{Zorzi_2022_CVPR}.

\begin{figure*}[t]
	\includegraphics[width=\textwidth]{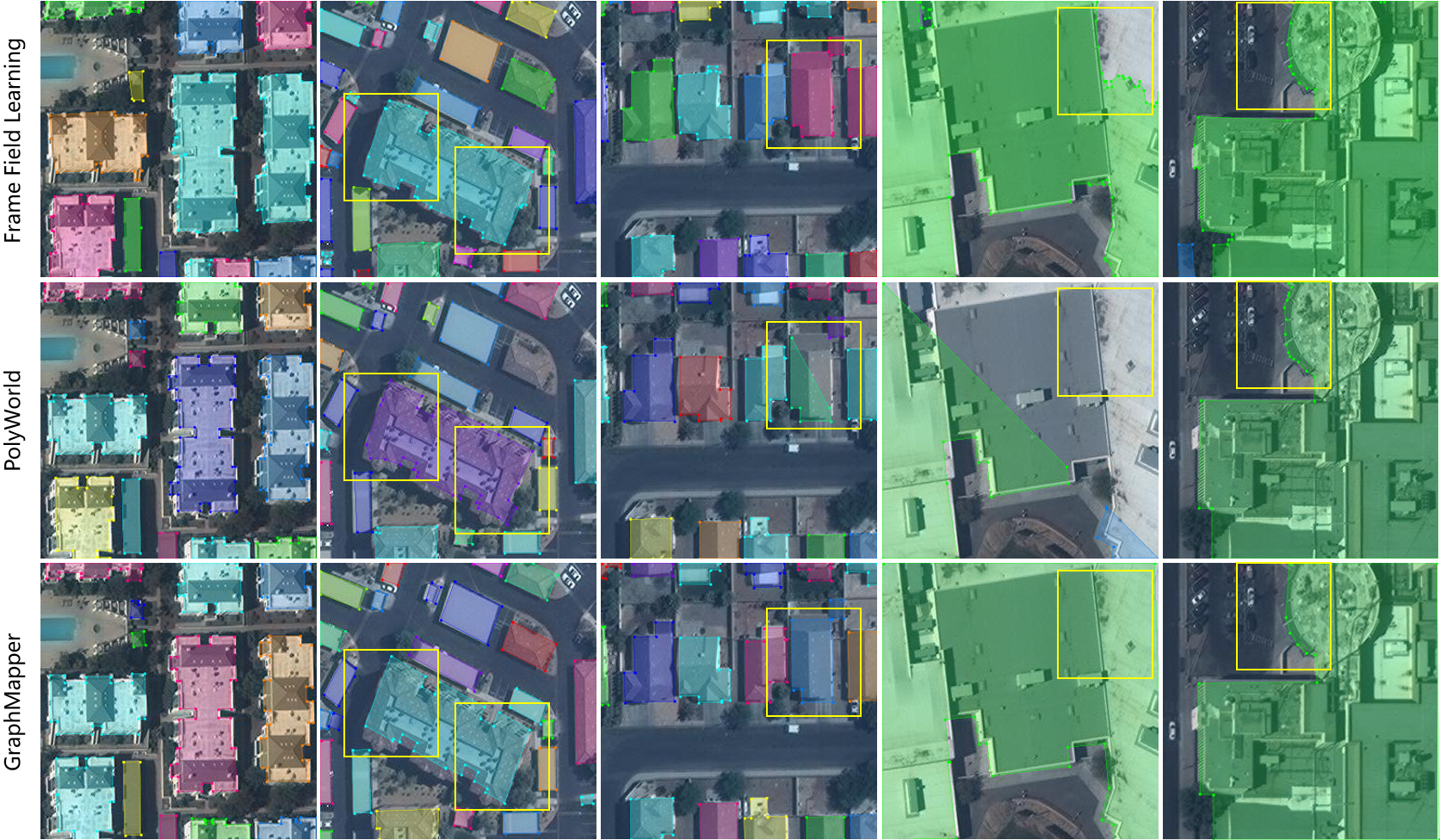}
	\centering
	\caption{Qualitative evaluation of building footprint mapping results. From top row to bottom row are results from Frame Field Learning \cite{Girard_2021_CVPR}, Polyworld \cite{Zorzi_2022_CVPR} and GraphMapper. Yellow boxes highlight the differences of different methods.}
	\label{fig:building_main}
\end{figure*}

\begin{table*}[th]
\small
\centering

\begin{tabular}{@{}l|llllll|llllll@{}}

\toprule
\textbf{Method}                   & $AP$   & $AP_{50}$ & $AP_{75}$ & $AP_{S}$  & $AP_M$  & $AP_L$  & $AR$   & $AR_{50}$ & $AR_{75}$ & $AR_S$  & $AR_M$  & $AR_L$  \\ \midrule
FFL (with field), mask       & 57.7 & 83.8 & 66.3 & 33.8 & 73.8 & 81.0 & 68.1 & 91.0 & 77.7 & 47.5 & 80.0 & 86.7 \\
FFL (with field), simple poly & 61.7 & 87.6 & 71.4 & 35.7 & 74.9 & 83.0 & 65.4 & 89.8 & 74.6 & 42.5 & 78.6 & 85.8 \\
FFL (with field), ACM poly    & 61.3 & 87.4 & 70.6 & 33.9 & 75.1 & 83.1 & 64.9 & 89.4 & 73.9 & 41.2 & 78.7 & 85.9 \\ \midrule
PolyWorld (offset off)     & 58.7 & 86.9 & 64.5 & 31.8 & 80.1 & 85.9 & 71.7 & 92.6 & 79.9 & 47.4 & 85.7 & 94.0 \\
PolyWorld (offset on)      & 63.3 & 88.6 & 70.5 & 37.2 & 83.6 & 87.7 & 75.4 & \textbf{93.5} & 83.1 & 52.5 & 88.7 & 95.2 \\ \midrule
GraphMapper, mask & 63.6 & 88.3 & 69.6 & 35.6 & 85.8 & 93.9 & 75.9 & 93.1 & 82.8 & 50.7 & 90.2 & 98.1 \\
GraphMapper, final & \textbf{72.8} & \textbf{89.1} & \textbf{79.7} & \textbf{46.6} & \textbf{90.6} & \textbf{91.3} & \textbf{83.1} & 93.3 & \textbf{88.1} & \textbf{61.5} & \textbf{95.2} & \textbf{97.2} \\
 \bottomrule

\end{tabular}

\caption{COCO evaluation results for building on CrowdAI Dataset.}
\label{tab:building_main}
\end{table*}

\begin{table}[htb]
\small
\centering
\begin{tabular}{l|c@{\hspace{1.2\tabcolsep}} c@{\hspace{1.2\tabcolsep}} c@{\hspace{1.2\tabcolsep}} c}
\toprule
\textbf{Method}   &   IoU & C-IoU & MTA & N ratio \\
\midrule
FFL (no field), simple poly &  83.9 & 23.6  & 51.8°  & 5.96 \\
FFL (with field), simple poly & 84.0 & 30.1 & 48.2° & 2.31 \\
FFL (with field), ACM poly & 84.1 & 73.7 & 33.5° & 1.13 \\
\midrule
PolyWorld (offset off) & 89.9 & 86.9 & 35.0° & 0.93 \\
PolyWorld (offset on) & 91.3 & 88.2 & 32.9° & 0.93 \\
\midrule
GraphMapper & \boldmath$93.9$ & \boldmath$88.8$ & \boldmath$30.4^{\circ}$ & 1.01\\
\bottomrule
\end{tabular}

\caption{IoU, MTA, C-IoU and N ratio evaluation results on CrowdAI Dataset.}
\label{tab:building_iou_mta}
\end{table}

\subsection{Benchmark results}
\textbf{Road network extraction.} 
We compare GraphMapper with Sat2Graph \cite{He_2020_ECCV} in Fig. \ref{fig:road_main}, which is considered the state-of-the-art method in road network mapping. GraphMapper can generate visually comparable results to Sat2Graph, but with less redundant roads in both datasets. Also, GraphMapper doesn't require tedious post-processing. 

Quantitatively, GraphMapper consistently shows improvement for APLS and TOPO on both road datasets (Tab. \ref{tab:road_main}) compared to Sat2Graph \cite{He_2020_ECCV}, where TOPO F1 is improved by $6.2\sim 9.3$, and APLS is improved by $4.4\sim 4.9$.


\begin{table*}[th]
\small
\centering
	\begin{tabular}{@{}c|llll|llll@{}}
	\toprule
	\multirow{2}{*}{\textbf{Method}}         & \multicolumn{4}{c|}{City-Scale Dataset}                  & \multicolumn{4}{c}{SpaceNet Roads Dataset} \\
									& $Prec.\uparrow$                & $Rec.\uparrow$                  & $F_1\uparrow$ & $APLS\uparrow$ & $Prec.\uparrow$  & $Rec.\uparrow$   & $F_1\uparrow$  & $APLS\uparrow$ \\
	\midrule
	RoadTracer\cite{Bastani_2018_CVPR} & 78.0  &  57.4 & 66.2 & 57.3 & 78.6 & 62.5 &  69.6 & 56.0 \\	
    \multicolumn{1}{l|}{Sat2Graph\cite{He_2020_ECCV}} & \multicolumn{1}{l}{80.7} & \multicolumn{1}{l}{72.3} & 76.3 & 63.1 & 85.9 & \multicolumn{1}{l}{ 76.6} & 80.9 & 64.4 \\
	\midrule
	\multicolumn{1}{l|}{GraphMapper}   & \multicolumn{1}{l}{\boldmath$89.6$} & \multicolumn{1}{l}{\boldmath$82.6$} &  \boldmath$85.6$  & \boldmath$68.0$ &  \boldmath$90.7$   & \multicolumn{1}{l}{\boldmath$84.6$}  & \boldmath$87.1$   &  \boldmath$68.8$    \\
	\bottomrule
	\end{tabular}

\caption{Comparison of road TOPO and APLS evaluation metric.}
\label{tab:road_main}
\end{table*}

\begin{table}[thb]
\small
\centering
\begin{tabular}{l|c@{\hspace{1.2\tabcolsep}} c@{\hspace{1.2\tabcolsep}} c@{\hspace{1.2\tabcolsep}} c}
\toprule
\textbf{Method}   &   IoU & C-IoU & MTA & N ratio \\
\midrule
simple poly & 92.3 & 81.6 &  38.5° & 1.26 \\
w/o incremental & 93.8 & 78.3 & 34.6°  & 1.11 \\
w/o reg & \boldmath$94.4$ & 82.7 & 31.5° & 1.28\\
\midrule
GraphMapper & 94.3 & \boldmath$89.5$ & \boldmath$30.4$° & \boldmath$1.01$\\
\bottomrule
\end{tabular}
\caption{Building ablation study on CrowdAI Dataset.}
\label{tab:building_ablation}
\end{table}

\begin{table}[thb]
\small
\centering
\begin{tabular}{l|l@{\hspace{1.5\tabcolsep}} l@{\hspace{1.5\tabcolsep}} l@{\hspace{1.5\tabcolsep}} l}
	\toprule
	\textbf{Method}     &   $Pre.$ & $Rec.$ & $F_1$ & $APLS$ \\
	\midrule 
	w/o sort	& 87.2 & 80.1 & 82.5 & 66.2 \\
	w/o incremental & 89.6 &  81.2 & 84.8 & 67.6\\
	GraphMapper & \boldmath$89.9$ & \boldmath$82.9$ & \boldmath$85.9$& \boldmath$68.9$   \\
	\bottomrule
\end{tabular}

\caption{Road Ablation study on City-Scale dataset.}
\label{tab:road_ablation}
\end{table}

\textbf{Building footprint extraction.} 
Qualitative evaluation results are reported in Fig. \ref{fig:building_main}. We compare GraphMapper with two recent methods, Frame Field Learning (FFL) \cite{Girard_2021_CVPR} and PolyWorld \cite{Zorzi_2022_CVPR}, that represent the state-of-the-art in building polygon mapping. All methods can accurately capture small and simple building polygons. 
GraphMapper shows the best shape regularization, polygon simplicity and accuracy.

Quantitative evaluation results are reported in Tab. \ref{tab:building_main}. GraphMapper significantly outperforms Polyworld \cite{Zorzi_2022_CVPR} by 9.5/7.7 in COCO AP/AR.
The improvements are largely contributed by primitive refinement and relationship reconstruction, which improved AP/AR by 9.2/7.2 to segmentation mask (\emph{GraphMapper, mask}). We report IoU, C-IoU and MTA evaluation results in Tab. \ref{tab:building_iou_mta}. GraphMapper outperform all competing methods in all metrics. The improved C-IoU and MTA numerically demonstrated that GraphMapper can generate more regularized vector shapes compared to previous methods, which is consistent with our visual comparison. Our N ratio is also closer to 1 compared to other methods, which suggest our method generate polygons with more similar complexity to ground truth.

\subsection{Ablation Study and Discussion}
We report ablation study results in Tab. \ref{tab:building_ablation} for building and Tab. \ref{tab:road_ablation} for road. In these two tables, \emph{GraphMapper} is our proposed model; \emph{simple poly} uses DP algorithm \cite{douglas_1973} over traced polygons without primitive refinement or relationship reconstruction; \emph{w/o incremental} refines primitives and predicts relationships in parallel with shared MHA encoder; \emph{w/o reg} removes regularization loss $L_{reg}$ during training; 
\emph{w/o sort} classifies relationship by thresholding relationship probability in $E$.

\textbf{Incremental Reconstruction.} For building mapping, incremental modeling (\emph{GraphMapper}) shows to significantly improve C-IoU by 11.2\%  and MTA by 4.2\% compared to parallel modeling (\emph{w/o incremental}). Similarly for road mapping, TOPO and ALPS are improved with incremental reconstruction. We believe that improved performance of incremental modeling is caused by the improved accuracy of ground truth matching for relationship learning when using refined primitives instead of initial primitives. 

\textbf{Shape regularization.} Relationship reconstruction for shape regularization shows to improve C-IoU by 6.8 and MTA by 1.1° compared to without shape regularization loss (\emph{w/o reg}) for building mapping, which suggests that relationship learning with consistency between primitives' properties and their relationships can effectively regularize shapes for vector mapping. 

\textbf{Sorting in embedding space.} We compare out point connections strategy with naive connectivity relationship classification (\emph{w/o sort}) in road mapping. Our method is shown to improve TOPO F1 by 3.4 and ALPS by 2.7 compared to standard relationship classification (\emph{w/o sort}). We found the improvement is mainly due to reduced redundant connections between points, and reduced sensitivity to the threshold (See supplementary materials) of connectivity probability. 



\section{Conclusions}
We propose GraphMapper, an end-to-end model for unified vector mapping from satellite images. By converting vector mapping tasks into primitive graph estimation tasks, GraphMapper can explicitly model topology reconstruction and shape regularization without tedious post-processing. We applied GraphMapper to building and road mapping with few task-specific designs and achieved favorable performance to existing methods. The simplicity and strong performance of GraphMapper effectively reduced the complexity for comprehensive mapping tasks. 

There are several directions for future work. One direction is to further improve the compatibility between shape learning features and image recognition features for better co-learning of vector shapes and image features. Another direction is that GraphMapper relies on primitive sampling to capture geometric critical locations, which can be challenging for complicated scenarios, such as junctions and neighboring parallel roads. Predict point sequences along road can reduce the need for point sampling, but unable to use global map geometry context for optimal  decision making. How to combine the benefits of these two methods can be an interesting future research direction.

\clearpage
{\small
\bibliographystyle{ieee_fullname}
\bibliography{references}

\begin{thebibliography}{10}\itemsep=-1pt

\bibitem{MSBuilding}
Microsoft {US} building footprints.
\newblock \url{https://github.com/microsoft/USBuildingFootprints}.
\newblock Accessed: 2018-07-14.

\bibitem{Acuna_2018_CVPR}
David Acuna, Huan Ling, Amlan Kar, and Sanja Fidler.
\newblock Efficient interactive annotation of segmentation datasets with
  polygon-rnn++.
\newblock In {\em Proceedings of the IEEE Conference on Computer Vision and
  Pattern Recognition (CVPR)}, June 2018.

\bibitem{Bardes2021}
Adrien Bardes, Jean Ponce, and Yann LeCun.
\newblock Vicreg: Variance-invariance-covariance regularization for
  self-supervised learning, 2021.

\bibitem{Bastani_2018_CVPR}
Favyen Bastani, Songtao He, Sofiane Abbar, Mohammad Alizadeh, Hari
  Balakrishnan, Sanjay Chawla, Sam Madden, and David DeWitt.
\newblock Roadtracer: Automatic extraction of road networks from aerial images.
\newblock In {\em Proceedings of the IEEE Conference on Computer Vision and
  Pattern Recognition (CVPR)}, June 2018.

\bibitem{Batra_2019_CVPR}
Anil Batra, Suriya Singh, Guan Pang, Saikat Basu, C.V. Jawahar, and Manohar
  Paluri.
\newblock Improved road connectivity by joint learning of orientation and
  segmentation.
\newblock In {\em Proceedings of the IEEE/CVF Conference on Computer Vision and
  Pattern Recognition (CVPR)}, June 2019.

\bibitem{berman2018lovasz}
Maxim Berman, Amal~Rannen Triki, and Matthew~B Blaschko.
\newblock The lov{\'a}sz-softmax loss: A tractable surrogate for the
  optimization of the intersection-over-union measure in neural networks.
\newblock In {\em Proceedings of the IEEE conference on computer vision and
  pattern recognition}, pages 4413--4421, 2018.

\bibitem{Biagioni_2012_TRR}
James Biagioni and Jakob Eriksson.
\newblock Inferring road maps from global positioning system traces: Survey and
  comparative evaluation.
\newblock {\em Transportation Research Record}, 2291(1):61--71, 2012.

\bibitem{carion_2020_ECCV}
Nicolas Carion, Francisco Massa, Gabriel Synnaeve, Nicolas Usunier, Alexander
  Kirillov, and Sergey Zagoruyko.
\newblock End-to-end object detection with transformers.
\newblock In Andrea Vedaldi, Horst Bischof, Thomas Brox, and Jan-Michael Frahm,
  editors, {\em Computer Vision -- ECCV 2020}, pages 213--229, Cham, 2020.
  Springer International Publishing.

\bibitem{Castrejon_2017_CVPR}
Lluis Castrejon, Kaustav Kundu, Raquel Urtasun, and Sanja Fidler.
\newblock Annotating object instances with a polygon-rnn.
\newblock In {\em Proceedings of the IEEE Conference on Computer Vision and
  Pattern Recognition (CVPR)}, July 2017.

\bibitem{CHEN2020114}
Qi Chen, Lei Wang, Steven~L. Waslander, and Xiuguo Liu.
\newblock An end-to-end shape modeling framework for vectorized building
  outline generation from aerial images.
\newblock {\em ISPRS Journal of Photogrammetry and Remote Sensing},
  170:114--126, 2020.

\bibitem{Chen_2021_CVPR}
Xinlei Chen and Kaiming He.
\newblock Exploring simple siamese representation learning.
\newblock In {\em Proceedings of the IEEE/CVF Conference on Computer Vision and
  Pattern Recognition (CVPR)}, pages 15750--15758, June 2021.

\bibitem{Cheng_2019_CVPR}
Dominic Cheng, Renjie Liao, Sanja Fidler, and Raquel Urtasun.
\newblock Darnet: Deep active ray network for building segmentation.
\newblock In {\em Proceedings of the IEEE/CVF Conference on Computer Vision and
  Pattern Recognition (CVPR)}, June 2019.

\bibitem{douglas_1973}
David~H Douglas and Thomas~K Peucker.
\newblock Algorithms for the reduction of the number of points required to
  represent a digitized line or its caricature.
\newblock {\em Cartographica: The International Journal for Geographic
  Information and Geovisualization}, 10(2):112--122, 1973.

\bibitem{Etten_2020_WACV}
Adam~Van Etten.
\newblock City-scale road extraction from satellite imagery v2: Road speeds and
  travel times.
\newblock In {\em Proceedings of the IEEE/CVF Winter Conference on Applications
  of Computer Vision (WACV)}, March 2020.

\bibitem{vanetten2019spacenet}
Adam~Van Etten, Dave Lindenbaum, and Todd~M. Bacastow.
\newblock Spacenet: A remote sensing dataset and challenge series, 2019.

\bibitem{Girard_2021_CVPR}
Nicolas Girard, Dmitriy Smirnov, Justin Solomon, and Yuliya Tarabalka.
\newblock Polygonal building extraction by frame field learning.
\newblock In {\em Proceedings of the IEEE/CVF Conference on Computer Vision and
  Pattern Recognition (CVPR)}, pages 5891--5900, June 2021.

\bibitem{NEURIPS2020_f3ada80d}
Jean-Bastien Grill, Florian Strub, Florent Altch\'{e}, Corentin Tallec, Pierre
  Richemond, Elena Buchatskaya, Carl Doersch, Bernardo Avila~Pires, Zhaohan
  Guo, Mohammad Gheshlaghi~Azar, Bilal Piot, koray kavukcuoglu, Remi Munos, and
  Michal Valko.
\newblock Bootstrap your own latent - a new approach to self-supervised
  learning.
\newblock In H. Larochelle, M. Ranzato, R. Hadsell, M.F. Balcan, and H. Lin,
  editors, {\em Advances in Neural Information Processing Systems}, volume~33,
  pages 21271--21284. Curran Associates, Inc., 2020.

\bibitem{Mordechai_2008}
Mordechai Haklay and Patrick Weber.
\newblock Openstreetmap: User-generated street maps.
\newblock {\em IEEE Pervasive Computing}, 7(4):12--18, 2008.

\bibitem{He_2020_ECCV}
Songtao He, Favyen Bastani, Satvat Jagwani, Mohammad Alizadeh, Hari
  Balakrishnan, Sanjay Chawla, Mohamed~M. Elshrif, Samuel Madden, and
  Mohammad~Amin Sadeghi.
\newblock Sat2graph: Road graph extraction through graph-tensor encoding.
\newblock In Andrea Vedaldi, Horst Bischof, Thomas Brox, and Jan-Michael Frahm,
  editors, {\em Computer Vision -- ECCV 2020}, pages 51--67, Cham, 2020.
  Springer International Publishing.

\bibitem{Kass_1988}
Michael Kass, Andrew Witkin, and Demetri Terzopoulos.
\newblock Snakes: Active contour models.
\newblock {\em International Journal of Computer Vision}, 1:321–331, 1988.

\bibitem{adam}
Diederik~P. Kingma and Jimmy Ba.
\newblock Adam: A method for stochastic optimization, 2014.

\bibitem{Li_2019_ICCV}
Zuoyue Li, Jan~Dirk Wegner, and Aurelien Lucchi.
\newblock Topological map extraction from overhead images.
\newblock In {\em Proceedings of the IEEE/CVF International Conference on
  Computer Vision (ICCV)}, October 2019.

\bibitem{Liang_2020_CVPR}
Justin Liang, Namdar Homayounfar, Wei-Chiu Ma, Yuwen Xiong, Rui Hu, and Raquel
  Urtasun.
\newblock Polytransform: Deep polygon transformer for instance segmentation.
\newblock In {\em Proceedings of the IEEE/CVF Conference on Computer Vision and
  Pattern Recognition (CVPR)}, June 2020.

\bibitem{Marcos_2018_CVPR}
Diego Marcos, Devis Tuia, Benjamin Kellenberger, Lisa Zhang, Min Bai, Renjie
  Liao, and Raquel Urtasun.
\newblock Learning deep structured active contours end-to-end.
\newblock In {\em Proceedings of the IEEE Conference on Computer Vision and
  Pattern Recognition (CVPR)}, June 2018.

\bibitem{Mattyus_2017_ICCV}
Gellert Mattyus, Wenjie Luo, and Raquel Urtasun.
\newblock Deeproadmapper: Extracting road topology from aerial images.
\newblock In {\em Proceedings of the IEEE International Conference on Computer
  Vision (ICCV)}, Oct 2017.

\bibitem{Mnih_2010}
Volodymyr Mnih and Geoffrey~E. Hinton.
\newblock Learning to detect roads in high-resolution aerial images.
\newblock In Kostas Daniilidis, Petros Maragos, and Nikos Paragios, editors,
  {\em Computer Vision -- ECCV 2010}, pages 210--223, Berlin, Heidelberg, 2010.
  Springer Berlin Heidelberg.

\bibitem{mohanty2020deep}
Sharada~Prasanna Mohanty, Jakub Czakon, Kamil~A Kaczmarek, Andrzej Pyskir,
  Piotr Tarasiewicz, Saket Kunwar, Janick Rohrbach, Dave Luo, Manjunath Prasad,
  Sascha Fleer, et~al.
\newblock Deep learning for understanding satellite imagery: An experimental
  survey.
\newblock {\em Frontiers in Artificial Intelligence}, 3, 2020.

\bibitem{Mosinska_2018_CVPR}
Agata Mosinska, Pablo Márquez-Neila, Mateusz Koziński, and Pascal Fua.
\newblock Beyond the pixel-wise loss for topology-aware delineation.
\newblock In {\em Proceedings of the IEEE Conference on Computer Vision and
  Pattern Recognition (CVPR)}, June 2018.

\bibitem{Nauata_2020_ECCV}
Nelson Nauata and Yasutaka Furukawa.
\newblock Vectorizing world buildings: Planar graph reconstruction by primitive
  detection and relationship inference.
\newblock In Andrea Vedaldi, Horst Bischof, Thomas Brox, and Jan-Michael Frahm,
  editors, {\em Computer Vision -- ECCV 2020}, pages 711--726, Cham, 2020.
  Springer International Publishing.

\bibitem{sirko2021continentalscale}
Wojciech Sirko, Sergii Kashubin, Marvin Ritter, Abigail Annkah, Yasser
  Salah~Eddine Bouchareb, Yann Dauphin, Daniel Keysers, Maxim Neumann,
  Moustapha Cisse, and John Quinn.
\newblock Continental-scale building detection from high resolution satellite
  imagery, 2021.

\bibitem{sudre_2017}
Carole~H. Sudre, Wenqi Li, Tom Vercauteren, Sebastien Ourselin, and M.
  Jorge~Cardoso.
\newblock Generalised dice overlap as a deep learning loss function for highly
  unbalanced segmentations.
\newblock In M.~Jorge Cardoso, Tal Arbel, Gustavo Carneiro, Tanveer
  Syeda-Mahmood, Jo{\~a}o Manuel~R.S. Tavares, Mehdi Moradi, Andrew Bradley,
  Hayit Greenspan, Jo{\~a}o~Paulo Papa, Anant Madabhushi, Jacinto~C.
  Nascimento, Jaime~S. Cardoso, Vasileios Belagiannis, and Zhi Lu, editors,
  {\em Deep Learning in Medical Image Analysis and Multimodal Learning for
  Clinical Decision Support}, pages 240--248, Cham, 2017. Springer
  International Publishing.

\bibitem{Tan_2020_CVPR}
Yong-Qiang Tan, Shang-Hua Gao, Xuan-Yi Li, Ming-Ming Cheng, and Bo Ren.
\newblock Vecroad: Point-based iterative graph exploration for road graphs
  extraction.
\newblock In {\em Proceedings of the IEEE/CVF Conference on Computer Vision and
  Pattern Recognition (CVPR)}, June 2020.

\bibitem{Vaswani_2017_NIPS}
Ashish Vaswani, Noam Shazeer, Niki Parmar, Jakob Uszkoreit, Llion Jones,
  Aidan~N Gomez, \L~ukasz Kaiser, and Illia Polosukhin.
\newblock Attention is all you need.
\newblock In I. Guyon, U.~Von Luxburg, S. Bengio, H. Wallach, R. Fergus, S.
  Vishwanathan, and R. Garnett, editors, {\em Advances in Neural Information
  Processing Systems}, volume~30. Curran Associates, Inc., 2017.

\bibitem{Zhang_2020_CVPR}
Fuyang Zhang, Nelson Nauata, and Yasutaka Furukawa.
\newblock Conv-mpn: Convolutional message passing neural network for structured
  outdoor architecture reconstruction.
\newblock In {\em Proceedings of the IEEE/CVF Conference on Computer Vision and
  Pattern Recognition (CVPR)}, June 2020.

\bibitem{Zhou_2018_CVPR_Workshops}
Lichen Zhou, Chuang Zhang, and Ming Wu.
\newblock D-linknet: Linknet with pretrained encoder and dilated convolution
  for high resolution satellite imagery road extraction.
\newblock In {\em Proceedings of the IEEE Conference on Computer Vision and
  Pattern Recognition (CVPR) Workshops}, June 2018.

\bibitem{Zhou_2019_CVPR}
Yi Zhou, Connelly Barnes, Jingwan Lu, Jimei Yang, and Hao Li.
\newblock On the continuity of rotation representations in neural networks.
\newblock In {\em Proceedings of the IEEE/CVF Conference on Computer Vision and
  Pattern Recognition (CVPR)}, June 2019.

\bibitem{Zhou_2019_ICCV}
Yichao Zhou, Haozhi Qi, and Yi Ma.
\newblock End-to-end wireframe parsing.
\newblock In {\em Proceedings of the IEEE/CVF International Conference on
  Computer Vision (ICCV)}, October 2019.

\bibitem{Zorzi_2022_CVPR}
Stefano Zorzi, Shabab Bazrafkan, Stefan Habenschuss, and Friedrich Fraundorfer.
\newblock Polyworld: Polygonal building extraction with graph neural networks
  in satellite images.
\newblock In {\em Proceedings of the IEEE/CVF Conference on Computer Vision and
  Pattern Recognition (CVPR)}, pages 1848--1857, June 2022.

\bibitem{Zorzi_2020_ICPR}
Stefano Zorzi, Ksenia Bittner, and Friedrich Fraundorfer.
\newblock Machine-learned regularization and polygonization of building
  segmentation masks.
\newblock In {\em 2020 25th International Conference on Pattern Recognition
  (ICPR)}, pages 3098--3105, 2021.

\end{thebibliography}
}

\end{document}